\pdfoutput=1
\documentclass[11pt]{article}

\usepackage[final]{acl}

\usepackage[most]{tcolorbox}
\usepackage{times}
\usepackage{latexsym}

\usepackage[T1]{fontenc}
\usepackage[utf8]{inputenc}

\usepackage{microtype}

\usepackage{inconsolata}

\usepackage{graphicx}
\usepackage{pdfpages}

\usepackage{booktabs}
\usepackage{amsmath}
\usepackage{amsfonts}
\usepackage{hyperref}
\usepackage{array}
\usepackage{longtable}
\usepackage[caption=false]{subfig}

\title{\textsc{BehaviorBox}: Automated Discovery of Fine-Grained Performance Differences Between Language Models}

\author{Lindia Tjuatja \and Graham Neubig \\
   Carnegie Mellon University \\
 \small{
   \textbf{Correspondence:} \href{mailto:email@domain}{lindiat@andrew.cmu.edu}
 }}

\begin{document}
\maketitle
\begin{abstract}
Language model evaluation is a daunting task: prompts are brittle, corpus-level perplexities are vague, and the choice of benchmarks are endless.
Finding examples that show meaningful, generalizable differences between two LMs is crucial to understanding where one model succeeds and another fails. Can this process be done automatically? In this work, we propose methodology for automated comparison of language models that uses performance-aware contextual embeddings to find fine-grained features of text where one LM outperforms another.
Our method, which we name \textsc{BehaviorBox}, extracts coherent features that demonstrate differences with respect to the ease of generation between two LMs. Specifically, \textsc{BehaviorBox} finds features that describe groups of words in fine-grained contexts, such as \textit{conditional `were' in the phrase `if you were'} and \textit{exclamation marks after emotional statements}, where one model outperforms another within a particular datatset. We apply \textsc{BehaviorBox} to compare models that vary in size, model family, and post-training, and enumerate insights into specific contexts that illustrate meaningful differences in performance which cannot be found by measures such as corpus-level perplexity alone.\footnote{Code for this work is available at \url{https://github.com/lindiatjuatja/BehaviorBox}.}
\end{abstract}

\label{sec:comparison}
\begin{figure*}[!h]
    \centering
    {\includegraphics[width=0.97\textwidth]{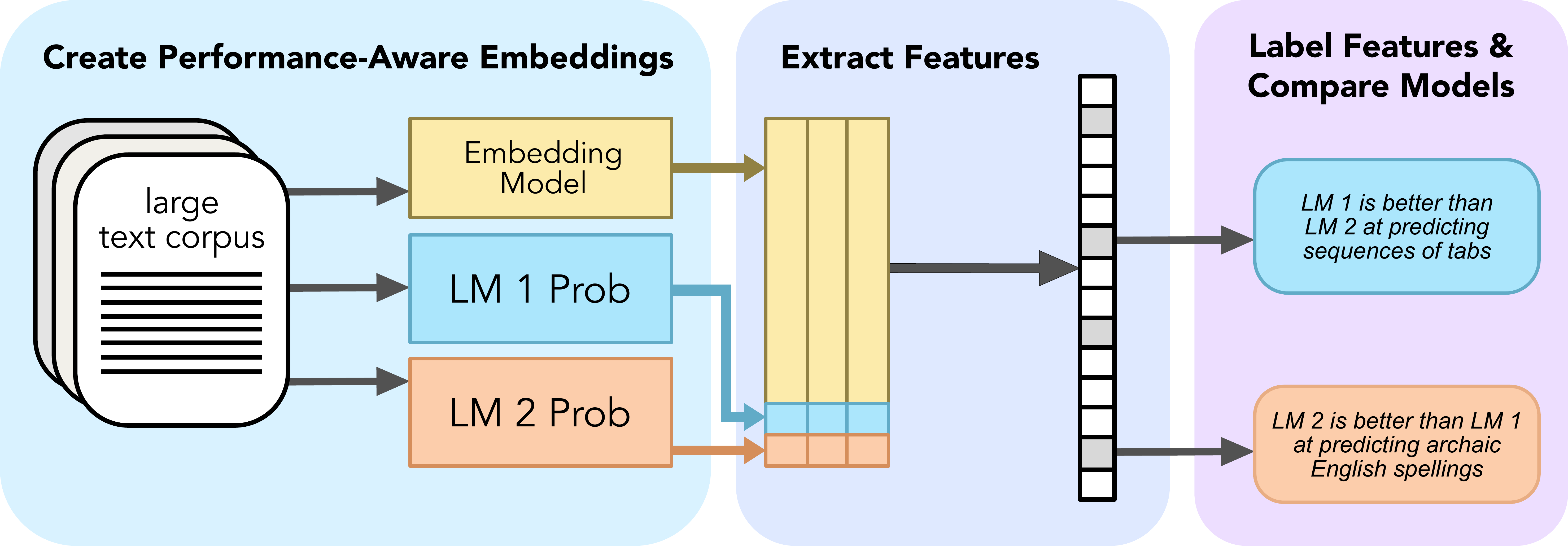}}
    \caption{\textsc{BehaviorBox} is a three-part automatic behavior comparison pipeline that discovers fine-grained features where one LM differs from another. These features are extracted from a corpus of \textit{performance-aware embeddings}, which take into account the semantics and usage of the text, along with measures of performance via probability under the evaluated LMs.}
    \label{fig:overview}
\end{figure*}

\section{Introduction}
\emph{Where does one language model perform better than another?} This deceptively simple question holds a near-endless number of complications. Practitioners must select from a dizzying array of evaluation methods, datasets, benchmarks, and metrics. Seemingly innocuous changes to evaluation pipelines, like the formatting of prompts, have been shown to drastically impact accuracy on a wide range of tasks \cite{sclarquantifying}. Even evaluating language models based on their original training objective---next token prediction---is not so straightforward. While metrics like perplexity \cite{jelinek1977perplexity} on a held-out corpus are commonly used and are generally correlated with downstream performance (e.g. \citealt{adiwardana2020towards,isik2024scaling}), the use of corpus-level perplexity on extremely large, diverse data often masks finer-grained differences on particular subgroups and domains~\cite{Magnusson2023PalomaAB, fang2024wrong}. 

What could an alternative to collections of benchmarks and corpus-level perplexity look like? One solution would be to partition the data into slices and report performance across these sub-corpora, as was done in Paloma~\cite{Magnusson2023PalomaAB}. However, such an approach depends on both knowing the relevant partitions ahead of time and having sufficient metadata such that these partitions can be made. But what if we could instead discover what the relevant features of these partitions are, and automatically generate a report telling practitioners specific and coherent groups of text where one model outperforms another?

We attempt to tackle this problem using our new evaluation method---\textsc{BehaviorBox}---which discovers fine-grained, human-interpretable features of data where one LM performs better than another. Unlike evaluations that depend on predetermined domains of data, \textsc{BehaviorBox} is a bottom-up approach that finds semantic and/or structural features of text where one model outperforms another, and does so independently of the domain or corpus the text originates from. As a consequence, \textsc{BehaviorBox} is capable of finding specific features and relationships in text that span across documents and domains, without the need to partition these domains ahead of time.

To find these features, \textsc{BehaviorBox} not only considers the \emph{context} of a text sample (via a contextual embedding), but also factors in the evaluated LMs' \emph{performance} on that sample (via the probabilities the models assign to the text), forming a performance-aware contextual representation of each text sample. After generating a large dataset of these representations, we then train a sparse autoencoder (SAE), which learns simple linear decompositions of the dense representations, with each component of the sparse representation acting as a discovered feature. Finally, using the groups of data determined by the SAE features, we generate natural language descriptions of each group.

We demonstrate the efficacy of \textsc{BehaviorBox} in discovering fine-grained differences between models in the language modeling task by comparing models that differ in size, family, and in types of post-training; specifically, we look at base and post-trained models of two sizes (7B and 13B parameters) across two model families, Llama 2 \cite{Touvron2023Llama2O} and OLMo 2 \cite{olmo20242}. Using \textsc{BehaviorBox}, we are able to find extremely fine-grained features in data that point to larger models' ability to better predict long-tailed stylized text (e.g. archaic spelling and vernacular), as well as show particular features related to dialogue and conversation where chat/RLHF-ed models excel. We are also able to discover differences between models that otherwise show near-identical performance with respect to perplexity, such as differences in predicting particular structure or formatting in text or different syntactic constructions in specific contexts. The insights provided by \textsc{BehaviorBox} provide a more holistic and detailed perspective on LM performance, and can be used to augment existing methods for evaluation and interpretability.

\section{Background}
\textsc{BehaviorBox} draws both conceptually and methodologically from two well-established areas of research: the problem of \emph{slice finding} and the \emph{behavioral evaluation} of black-box NLP systems.

\subsection{Slice Finding}
A key component in debugging and building better machine learning and NLP systems is identifying where and when a system underperforms. When we evaluate these systems, we may use overall metrics, such as accuracy on a benchmark or perplexity on a large corpus. However, overall performance may obfuscate stark differences in performance across subgroups; thus, if we are interested in the performance on groups within the larger dataset, we may partition the data into predetermined categories, and compare performance within these groups. Nevertheless, it is often difficult to know \emph{a priori} what the relevant groups of data are with respect to model performance. The task of automatically identifying salient groups of data where a model underperforms is known as \emph{slice finding} \cite{Chung2018SliceFA}, and is applicable across all sorts of tasks and modalities, from image classification to question answering.

Early works in slice finding often relied on metadata to find relevant slices \cite{Chung2018SliceFA}, but such an approach depends on the appropriate metadata categories to be specified and present in the data, which may not necessarily be the case. To solve this problem, slice finding methods such as George \cite{sohoni2020no}, Spotlight \cite{10.1145/3531146.3533240}, and Domino \cite{eyubogludomino} utilize learned representations of the data to find semantically similar clusters of underperforming samples. These methods have primarily focused on image classification tasks, and a few constrained natural language tasks, such as sentiment analysis.

\textsc{BehaviorBox} takes a similar approach as these works by utilizing contextual representations, but differs in two major ways. First, we focus on the language modeling task, which involves a significantly more complex output space compared to the tasks explored in prior work. Second, we focus on model \emph{comparison} as opposed to where a single model is ``incorrect'', as such a distinction is much less clear in the context of text generation.

\subsection{Behavioral Evaluation in NLP}
As NLP systems have grown ever more complex, efforts to better understand these largely black-box systems have become increasingly important. One approach to better understanding such a system is to generate explanations for a system's \emph{behaviors}, i.e. how a system's output changes when provided inputs of a certain type \cite{ribeiro-etal-2020-beyond}. Explanations usually take the form of a relationship between a particular feature in the data and the resulting prediction, e.g. the impact of the use of negation on the predictions of a sentiment analysis model. These explanations not only need to faithfully capture model behaviors, but should also be human interpretable \cite{10.1145/2939672.2939778, 10.5555/3295222.3295230}. 

In the context of explaining errors of NLP systems, works like Errudite \cite{wu-etal-2019-errudite} and CheckList \cite{ribeiro-etal-2020-beyond} provide frameworks for practitioners to stress-test models on precise hypotheses regarding the impact of specific features. Nevertheless, these hypotheses still need to be specified ahead of time. \textsc{BehaviorBox} can be seen as a complementary approach by serving as a form of hypothesis discovery, where such hypotheses can then be further explored in various other evaluation frameworks.

\section{Method Overview: \textsc{BehaviorBox}}                                                                                                                                        
As shown in~\autoref{fig:overview}, \textsc{BehaviorBox} is an automatic behavior comparison pipeline for language modeling, comprised of three parts: 
\begin{enumerate}
    \item \textbf{Data generation}, which consists of calculating contextual embeddings and aligning these embeddings with probabilities under LMs for the same text (\S\ref{sec:data-generation}),
    \item \textbf{Extracting features} that are coherent and capture similarities and differences regarding performance between models (\S\ref{sec:sae}), and
    \item \textbf{Labeling groups of data} indicating performance differences between LMs among the discovered data slices (\S\ref{sec:labeling}).
\end{enumerate}

\noindent The unit of data used in this method can, in theory, be as small as a token or as large as a document. However, in our experiments, we focus on characterizing performance at (roughly) the word level. We decided on this level of abstraction to balance both granularity and salience, as well as for engineering convenience. While tokens serve as the atomic unit of generation and are closest to the training objective, they may be less human-interpretable and are harder to work with when aligning the different tokenizers of the embedding model and various LMs. On the other hand, larger structures like phrases and sentences may be easier to categorize in terms of salient groups, but may be difficult to parse for certain types of documents commonly included in pretraining and evaluation (e.g. code, mathematical expressions, or other non-linguistic textual data), and could furthermore mask more granular trends that may be of interest.

\section{Data Generation}
\label{sec:data-generation}

Prior work has shown that incorporating learned representations of the input data along with a model's predictions and gold labels helps with identifying unlabeled classes of data where said model underperforms \cite{eyubogludomino, sohoni2020no}. Drawing from these works, \textsc{BehaviorBox} uses contextual embeddings to provide semantic information about each word, along with probabilities generated by the evaluated LMs, which serve as a measure of the LMs' performance. For contextual embeddings, we use the last hidden layer of Longformer \cite{beltagy2020longformer}. 
% To calculate LM probabilities, we host and query models via vLLM \cite{10.1145/3600006.3613165}.

As previously mentioned, we use \textsc{BehaviorBox} to slice our data (some arbitrary text dataset) into groups of words. Collecting and aligning contextual embeddings and probabilities per word across models that utilize different tokenization processes requires a number of engineering decisions, such as determining the boundaries of words, subsequently combining or splitting token log probabilities when necessary, and handling strings longer than the context window of different models. 

We use Longformer's pre-tokenizer, which largely splits on regular whitespace, as our method of determining word boundaries. To aggregate token representations within a single word, we average the embeddings of the constituent tokens. For probabilities, we multiply the probabilities of constituent tokens.
For instance, given a word that spans tokens $n$ to $m$ in a sequence $w = \{ t_n, t_{n+1}, \ldots, t_m \}$:

\begin{equation}
\mathbf{e}_w = \frac{1}{(m-n+1)}\sum_{j=n}^{m} \mathbf{e}_{t_j}, \quad \mathbf{e}_{t_j} \in \mathbb{R}^{768}
\end{equation}
\begin{equation}
p_w = \prod_{j=n}^{m}P(t_j|t_1,\ldots,t_{j-1})
\end{equation}

Each datapoint in the resulting dataset is a vector of dimension 770, where the first 768 dimensions are from the Longformer embedding and the last two are the probabilities of the language models being compared.

\begin{equation}
\mathbf{x}_w = \begin{bmatrix}\mathbf{e}_w \\ p_{w,1} \\ p_{w,2}\end{bmatrix} \in \mathbb{R}^{770}
\end{equation}

\section{Extracting Features}
\label{sec:sae}
Once we have a dataset of aligned words and probabilities for the LMs we wish to compare, we now have to find a way to extract and label fine-grained slices of data. This needs to be done in such a way that the slices are composed of \emph{coherent} sets of words in context and the labels adequately \emph{explain} the slice in a human-interpretable manner.

Previous works in automatic slice finding that incorporate learned representations have used various clustering algorithms such as k-means clustering \cite{sohoni2020no, 10.1145/3531146.3533240} and Gaussian mixture models \cite{eyubogludomino}. However, as opposed to finding (hard) partitions in the data, we want to find specific \emph{features} associated with text where one model performs better of worse than another. These features need not form a true mathematical partition of the entire corpus, but can instead be treated as linear decompositions of each text sample, where each word in context is comprised of some number of these features.

Finding simple, linear decompositions of otherwise complex representations is a problem in a wide variety of settings in NLP, such as creating more interpretable word embeddings \cite{faruqui-etal-2015-sparse}, generating hypotheses about variables of importance within text datasets \cite{movva2025sparse}, and---more recently---interpreting the internal states of transformer models \citep[][\emph{inter alia}]{Cunningham2023SparseAF, Lieberum2024GemmaSO, Gao2024ScalingAE}. We take a similar methodological approach to these works by using sparse autoencoders to extract features relevant to performance differences between two LMs. Using the SAE, we can then extract slices corresponding to each feature by finding the words whose representations that lead to the highest activation value of that feature.

\subsection{Sparse Autoencoder Training}
Recall that the features we are looking for ideally have the following characteristics: they should be coherent, fine-grained, and capture performance differences between models. Balancing each of these criteria inform our use of various hyperparameters and regularization choices.

The sparse autoencoder consists of an encoder and decoder: the encoder takes as input a vector $\mathbf{x}$, which is a concatenation of the contextual word embedding and LM probabilities, and creates a sparse representation $\mathbf{f}(\mathbf{x})$. The decoder then reconstructs the input (denoted as $\hat{\mathbf{x}}$) from this sparse representation. $\sigma(\cdot)$ denotes the activation function.
\begin{equation}
    \mathbf{f}(\mathbf{x}) = \sigma(\mathbf{W}_{enc}\mathbf{x} + \mathbf{b}_{enc})
\end{equation}
\begin{equation}
    \mathbf{\hat{x}} = \mathbf{f}(\mathbf{x})\mathbf{W}_{dec} + \mathbf{b}_{dec}
\end{equation}

\noindent For $\sigma(\cdot)$, we use RELU \cite{Agarap2018DeepLU} to ensure non-negative values, as we conceptually want our features to be additive. The weights of $\mathbf{W}_{enc}$, $\mathbf{b}_{enc}$, $\mathbf{W}_{dec}$, and $\mathbf{b}_{dec}$ are learned by minimizing the $\text{L}_2$ distance between the reconstruction $\mathbf{\hat{x}}$ and the original input $\mathbf{x}$, using AdamW \cite{Loshchilov2017DecoupledWD} as our optimizer.

\subsection{Enforcing Sparsity} While allowing us to create a faithful representation of the original input, the above setup does not constrain the autoencoder to be sparse. As a way to enforce sparsity, we apply a batch-wise top-k operation to the pre-RELU SAE hidden state \cite{Makhzani2013kSparseA, Gao2024ScalingAE, bussmann2024batchtopk}: for some value $k$, we flatten the batch (of size $N$), and zero out all activations that are not in the top $N \times k$ activations. This allows us to directly enforce $\mathbb{E}[\text{L}_0]$ at the batch level, as opposed to using a proxy such as adding an $\text{L}_1$ penalty to the loss \cite{bricken2023towards}. 

\small
\begin{equation}
\mathbf{f}_{sparse}(\mathbf{x}) = \text{BatchTopK}\Big(\sigma(\mathbf{W}_{enc}\mathbf{x} + \mathbf{b}_{enc}),\; k\Big)
\end{equation}
\normalsize

\subsection{Balancing Context and Performance Awareness} Including the probabilities in the input to the SAE on its own does not guarantee that the SAE will utilize that information. One reason why the SAE may not utilize probabilities is simply because these two features are overwhelmed by the large number of embedding features' contribution to the $\text{L}_2$ loss. Thus, we up-weigh the probability features so that the magnitude of the probability components make up 70\% of the total magnitude of the input.
% \footnote{We include additional details on selecting this hyperparameter in Appendix \ltedit{TODO}.}

\paragraph{Hyperparameters}
The dimension of the sparse representation we learn is 3000 with $k=50$. We include a table of all SAE training hyperparameters and additional training details in Appendix \ref{sec:hyperparams}.

\section{Processing and Labeling Features}
\subsection{Processing and Filtering SAE Features}
\label{sec:labeling}
After training, we now need to extract the slice of words associated with each feature of the SAE. We do this by taking the same dataset of words used to train the SAE and find the top 50 words that lead to the highest non-zero activation value for each feature in $\mathbf{f}_{sparse}(\mathbf{x})$. For features that have less than 50 non-zero activations, we consider all samples; we do not consider features with less than 10 non-zero activations.

Different features may have different ranges and distributions of non-zero activation values. As we don’t want to consider samples whose activations, though non-zero, are extremely small compared to the highest value, we set a “dynamic” cutoff by including samples whose activation values are either in the top 75\% or are greater than 25\% of the highest activation value.\footnote{While this choice is somewhat arbitrary, some sort of cutoff (whether in this manner or a similar method) was necessary to select groups of words to consider as representative of a feature in our analysis.} 
% Then, for each feature and associated words, we get the context of that word from the document it originated from and concatenate the preceding and following 10 words. 

However, not every feature is indicative of a significant and consistent performance difference between models. To exclude those that are not,  we only consider features that either show a median probability difference among the filtered samples greater than 0.1 \textit{or} a median log-probability difference greater than 1 (i.e. the ratio of probabilities is greater than $e$). We decided on thresholding based on probability as well as log-probability differences to capture a set of features that can show both large differences in magnitude along with large relative differences (even if the absolute magnitude of the probabilities is small, as would be the case for rarer long-tailed phenomena).\footnote{We include detailed explanation of this choice of thresholding strategy in Appendix \ref{sec:thresholding}.}

\subsection{Labeling Procedure}
As manually labeling every slice across multiple SAE runs would take a prohibitive amount of time, we partially automate this process by using a strong LLM (Claude 3.5 Sonnet, \citealt{TheC3}) as an annotator. For a given feature, we have a multi-step annotation procedure. First, we prompt the LLM annotator to determine if a group of words and their contexts form a coherent group, and if so to provide a label describing this group. We then have a second round of annotation to validate the label by feeding the same label with examples to the LLM annotator, asking it to either keep the original label if it is appropriate, provide a new one if the current label does not accurately describe the examples, or invalidate the feature if the group is not coherent. We include the prompts used in Appendix \ref{sec:prompts}.

While convenient, this automated labeling process is not perfect. There are two main failure cases in this process. First, the label may be slightly inaccurate. If the label is inaccurate but can be corrected, the authors manually rewrote the label. Otherwise, if the label cannot be corrected (e.g. cases where less than 10 of the examples is described by the label and/or the entire group of examples do not form a coherent grouping), we throw out the feature.

\label{sec:comparison}
% \begin{figure*}[!h]
%     \centering
%     {\includegraphics[width=1\textwidth]{figures/median_diff_histograms.pdf}}
%     \caption{Histograms showing the distribution of median probability differences between two models across features after the thresholding and labeling steps as described in \S\ref{sec:labeling}; each curve represents the better performing model for that subset of features. Numbers in parentheses indicate the total number of features associated with each model. Histograms for median log probability differences are included in \autoref{fig:logprob_diff}.}%
%     \label{fig:histograms}
% \end{figure*}

\section{Differentiating Model Performance with \textsc{BehaviorBox}}
Interpretability methods are notoriously hard to evaluate effectively \citep{lipton2018mythos,arora2022explain}, and thus in this work we follow previous work on slice finding  \citep{Chung2018SliceFA} and largely rely on qualitative inspection of the trends discovered by our method to demonstrate its utility.
Specifically, we use \textsc{BehaviorBox} to perform comparisons on language models across three axes of variation:

\begin{table}[t]
    \centering
    % \small
    \begin{tabular}{lccc} 
    \toprule
    Model & Perplexity $\downarrow$ & $\Delta$ \\ 
    \midrule
    Llama-7B & 9.856 & -- \\
    Llama-7B-Chat & 13.911 & $+$4.055 \\
    Llama-13B & 8.773 & $-$1.083 \\
    % Llama-13B-Chat & 11.386 & $+$2.613 \\
    \midrule
    OLMo-7B & 9.803 & -- \\
    OLMo-7B-DPO & 12.762 & $+$2.959 \\ 
    % OLMo-7B-SFT & 11.145 & $+$1.342 \\
    OLMo-13B & 8.756 & $-$1.047 \\
    % OLMo-13B-DPO & 9.567 & $+$0.811 \\
    % OLMo-13B-SFT & 9.067 & $+$0.311 \\
    \bottomrule
    \end{tabular}
    \caption{Perplexity per word for each of the models evaluated (lower is better). $\Delta$ indicates the change in perplexity from the 7B model within the same family.}
    \label{tab:ppl}
\end{table}
\begin{figure*}[!h]
    \centering
    \subfloat{
        {\includegraphics[width=1\textwidth]{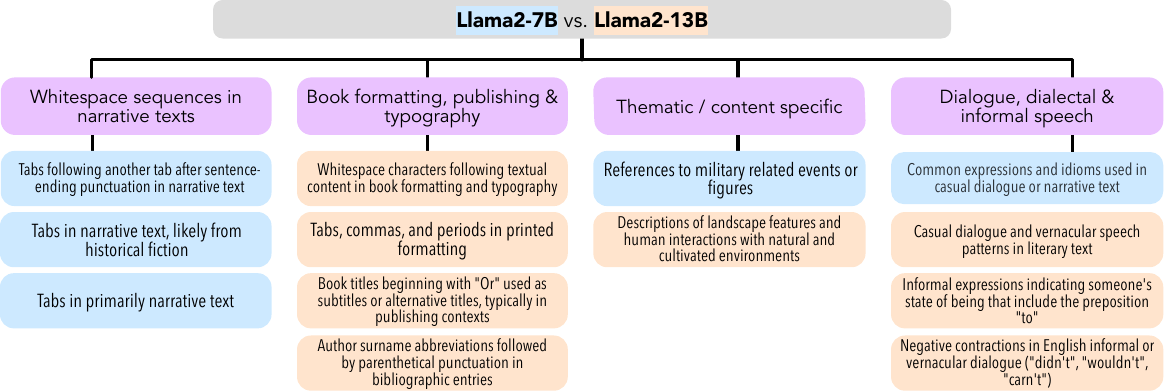}}
    } \\
    \subfloat{
        {\includegraphics[width=1\textwidth]{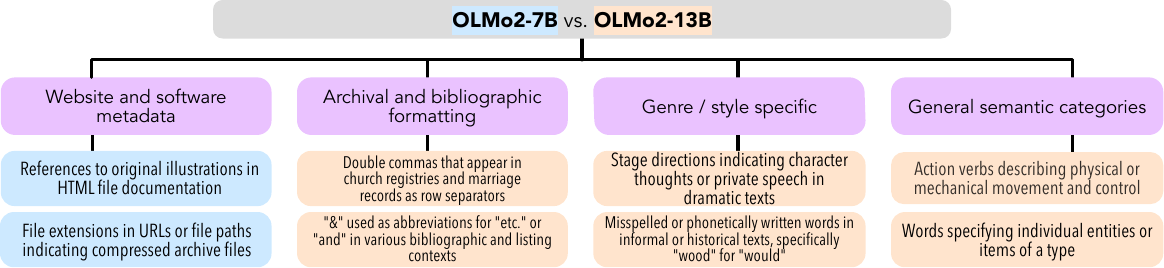}}
    }
    \caption{Representative features between 7B models (blue) and their 13B counterparts (orange).}%
    \label{fig:comp-size}
\end{figure*}

\begin{itemize}
\item \textbf{Model size:} 7B and 13B
\item \textbf{Post-training:} we compare base models against their post-trained RLHF-ed versions
\item \textbf{Model family:} Llama 2 \cite{Touvron2023Llama2O} and OLMo 2 \cite{olmo20242} (henceforth simply Llama and OLMo, respectively)
\end{itemize}

As the source of text we use to create the dataset of performance-aware representations, we use portions of the Dolma Dataset \cite{dolma}, an open dataset for language modeling containing a diverse mix of web content, academic publications, code, books, and encyclopedic materials. We sample 1000 documents across six of the data sources included in Dolma (Common Crawl, The Stack, C4, PeS2o, Project Gutenberg, and Wikipedia), totaling in approximately 80M words of data. As a baseline comparison between models, we include each model's perplexity per word on the subset of Dolma we use in our experiments in \autoref{tab:ppl}. In line with previous work, we find that the larger 13B base models have lower perplexity than their 7B counterparts \cite{kaplan2020scaling, xia-etal-2023-training}, while post-trained models that have undergone RLHF exhibit higher perplexity compared to their base counterparts \cite{Li2024PredictingVA}.

\paragraph{Interpreting Results} 
Recall that \textsc{BehaviorBox} generates features indicative of performance differences between the two models. While this on its own is interesting, these results are more easily understood when grouped into broader categories of features to make understanding these differences easier. 

\begin{figure*}[!h]
    \centering
    \subfloat{
        {\includegraphics[width=1\textwidth]{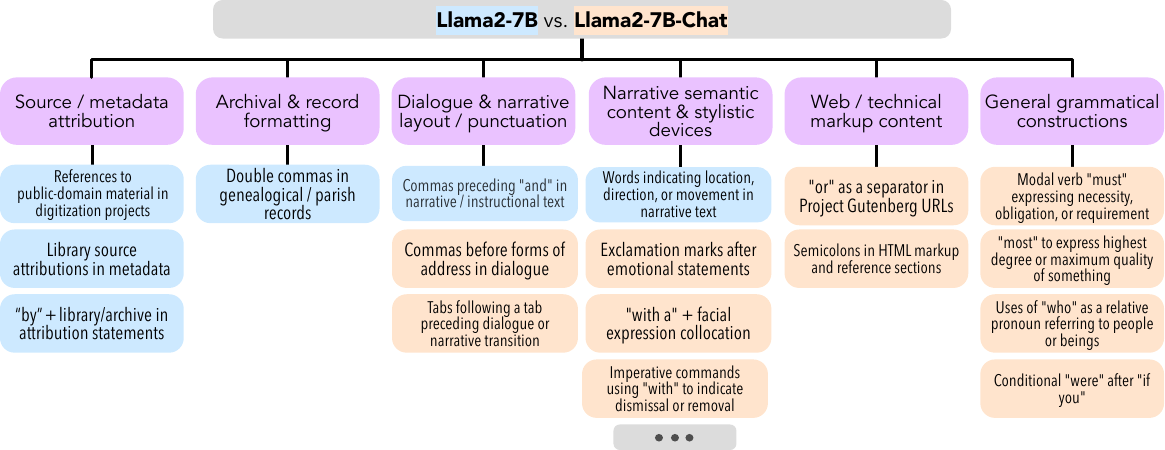}}
    } \\
    \subfloat{
        {\includegraphics[width=1\textwidth]{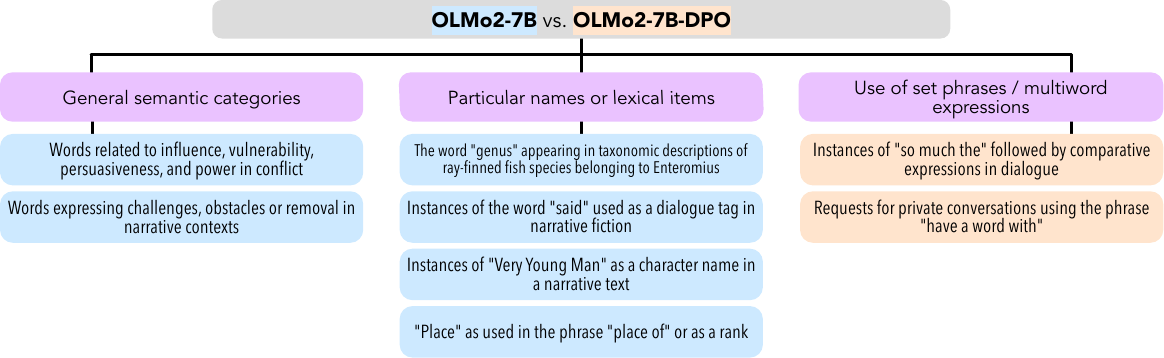}}
    }
    \caption{Representative features between base models (blue) and their post-trained counterparts (orange).}%
    \label{fig:comp-train}
\end{figure*}

\begin{figure*}[!h]
    \centering
    \subfloat{
        {\includegraphics[width=1\textwidth]{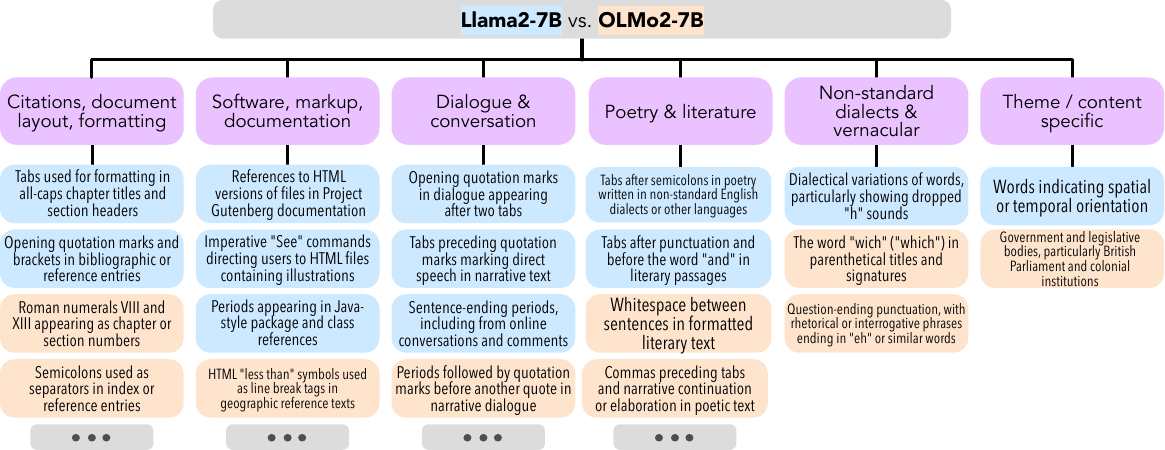}}
    } \\
    \subfloat{
        {\includegraphics[width=1\textwidth]{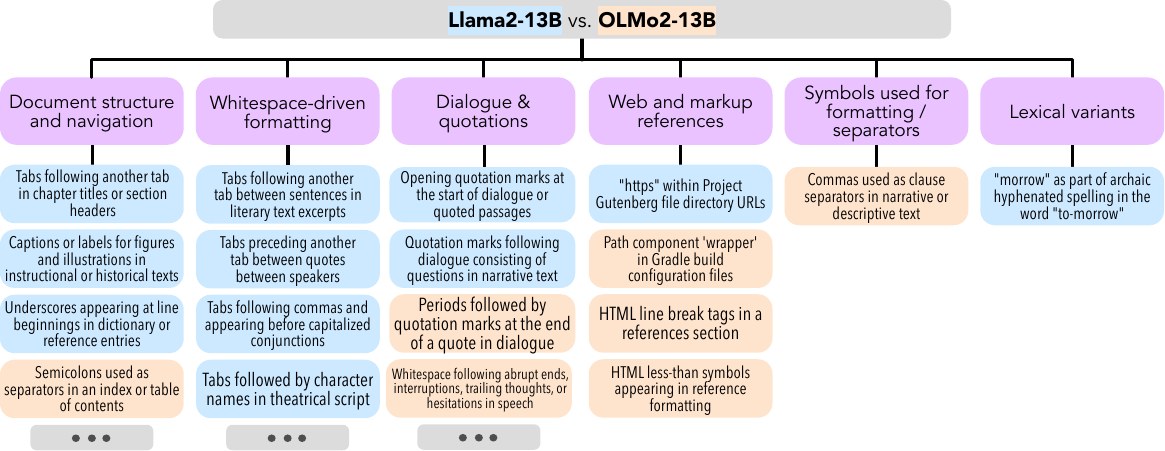}}
    }
    \caption{Representative features between Llama (blue) and OLMo models (orange) of the same sizes.}%
    \label{fig:comp-family}
\end{figure*}

In each of the below comparisons, we include dendrograms (see Figures \ref{fig:comp-size}, \ref{fig:comp-train}, and \ref{fig:comp-family}) which show these broad meta features (in purple) along with a representative sample of the features found by \textsc{BehaviorBox} (in blue and orange) within a single run.\footnote{A full list of features for each comparison can be found in Appendix \ref{sec:feature-labels}.} These meta features were found using a combination of data exploration and sense-making techniques \cite{chan2016comparing}, such as k-means clustering of the labels \cite{macqueen1967some} and prompting a LLM with the list of feature labels, along with manual inspection by the authors. 

As with many other forms of clustering, the data can be grouped into different sets depending on particular axes of interest; for example, we may be more interested in groups related to the actual \textit{form} of the word described by the feature labels, or we may be more interested in the overall \textit{context} the words occur in. In our visualizations, we mainly focus on the latter as similar surface forms can take on different meanings and functions depending on their context.

% Interestingly, we find that a greater difference in perplexity between a pair of models does not necessarily lead to a greater number of features discovered by \textsc{BehaviorBox}. \ltedit{This could indicate that...}

\subsection{Model Size}
In our first set of comparisons, we run \textsc{BehaviorBox} on Llama and OLMo models that vary in size. Both pairs of models have similar differences in perplexity, with larger models showing a $\approx1$ drop in perplexity compared to the 7B models; analogously, we find a greater number of features where 13B outperforms 7B for both Llama and OLMo, as shown in Figure \ref{fig:comp-size}.

For Llama-7B and 13B, we find that 7B outperforms 13B in whitespace sequences (specifically using tabs) within narrative texts, but 13B performs better at a wider range of formatting and typographic contexts, such as book titles and in bibliographic entries. Like Llama-13B, OLMo-13B also outperforms 7B with respect to particular aspects of archival and bibliographic formatting (use of double commas in a document of church registries, `\&' symbol in bibliographies and lists). Another overlapping aspect between the two size comparisons is that the larger models perform better on ``long-tail'' stylistic phenomena, such as vernacular and non-standard spellings. 

\subsection{Post-training}
Comparisons between base and post-trained models (the Chat version for Llama and DPO for OLMo) showed the highest differences in perplexity, with Llama-7B-Chat showing the greatest degradation in perplexity with over $+4$ gain, and nearly $+3$ gain for OLMo-7B-DPO from the base versions. While there is a larger difference in perplexity compared to the comparison between OLMo-7B and 13B, we don't find a greater number of features in this setting. Conversely, we find a greater number of features for the Llama2-7B and Llama2-7B-Chat comparison compared to 7B and 13B, though a majority of these features (17 out of 26) are actually for where the Chat version outperforms the base model, despite its significantly worse performance with respect to perplexity.

Between Llama-7B and Llama-7B-Chat, the base model performs better at text related to source/metadata attribution and archival formatting, while the chat version is better at a wider range of narrative stylistic features, markup and web related content, as well as particular grammatical constructions and phrases. For OLMo models, the base model is better at words related to categories like influence and obstacles, as well as particular instances of individual names or words (e.g. ``genus'' in descriptions of types of fish, ``Very Young Man'' as a character in a story), whereas the DPO version outperforms the base at predicting words in set phrases such as ``so much the X'' and ``have a word with Y''. Across both Llama and OLMo comparisons, the post-trained models appear to be better at conversational phrases and common multi-word grammatical constructions, which aligns with their intended use in chat settings.

\subsection{Model Families}
Figure \ref{fig:comp-family} visualizes the various features found by \textsc{BehaviorBox} between Llama and OLMo models same size. Despite the difference in perplexity being the lowest out of all comparisons ($\Delta\text{Perplexity} < 0.1$), we found the greatest number of distinct features between models of the same size between families. 

Furthermore, these features are persistent across size, as we can find sets of features that are shared among both 7B and 13B comparisons. OLMo models are better at predicting certain forms of separators like non-tab whitespace, semicolons, and commas, as well as uses of HTML in reference sections. On the other hand, common features for Llama models involve quotation marks, along with a substantial number of features involving sequences of tabs in various contexts such as document/text formatting and within prose. 

\section{\textsc{BehaviorBox} Features Distinguish Texts Generated by Different LMs}
\textsc{BehaviorBox} finds features based on data from a particular corpus. While this allows us to compare performance on controlled, naturalistic data, an open question remains of how the probabilities of predetermined strings translate to the model's behavior in open generation. Are features found by \textsc{BehaviorBox} also evident when comparing text generated from the compared models?

To investigate this, we look at whether the prevalence of particular strings mentioned in features found by \textsc{BehaviorBox} are more frequently generated by one model than another. We focus on comparing features and generations from Llama-13B and OLMo-13B, focusing on features that refer to a specific words for ease of analysis. For each model, we chose four strings based on the features where the model performs better: tab, quotation marks, ``https'', and ``morrow'' (in ``to-morrow'') for Llama-13B and periods followed by quotation marks, standard whitespace\footnote{A whitespace word when parsed by Longformer is actually due to two or more whitespaces being present in that position in the original document.}, HTML less than (``lt''), and commas for OLMo-13B.

For both models, we produce ``free-generations'' by conditioning only on the model's BOS token with a temperature of 1 and default max generation length \cite{Liu2025NotJustScaling}. We filter for generations that are between 400-600 words long, then among those sample 500 for each model. To test the hypothesis that a model produces a particular word significantly more frequently than another, we calculate the frequency of that string in every generation, then use the Mann-Whitney U test \cite{mann1947test} to see if the rates of generation of a string within a document is significantly more or less frequent for one model compared to the other.\footnote{We use the implementation from \url{https://docs.scipy.org/doc/scipy/reference/generated/scipy.stats.mannwhitneyu.html}, with ``greater'' or ``less'' for the alternative parameter.}

\begin{table}[t]
    \centering
    \footnotesize
    \begin{tabular}{lccc} 
    \toprule
    String & Hypothesis & p-value \\ 
    \midrule
    \textbf{tab} & L $>$ O & $4.8\text{e-}5$ \\
    \textbf{quotation mark} & L $>$ O & 0.02 \\ 
    \textbf{https} & L $>$ O & $8.5\text{e-}6$ \\
    morrow & L $>$ O & -- \\
    \midrule
    \textbf{period + quotation mark (.'')} & O $>$ L & 0.02 \\
    \textbf{whitespace} & O $>$ L & 0.03 \\
    HTML less than (lt) & O $>$ L & 0.16 \\
    \textbf{comma} & O $>$ L & $7.3\text{e-}4$ \\
    \bottomrule
    \end{tabular}
    \caption{Results from testing hypotheses about differences in string frequencies between Llama-13B and Olmo-13B; L $>$ O indicates that Llama-13B generates the string at a greater frequency than OLMo-13B and vice versa. \textbf{Bold} indicates significant results ($p < 0.05$).}
    \label{tab:freq}
\end{table}

As shown in Table \ref{tab:freq}, six of the eight string hypotheses lead to significant results, indicating that \textsc{BehaviorBox} features can be used to distinguish sets of generations between models. Additionally, while did not have a conclusive result for ``morrow'' (as it was not present in any of the sampled strings), we inspected 5000 free generations (not filtered for length) and found 20 occurrences of ``morrow'' (as used in ``to-morrow'') from Llama-13B and 0 from OLMo-13B, supporting our hypothesis. 

\section{Conclusion}
In this work, we introduced \textsc{BehaviorBox}, an automated pipeline for the behavioral comparison of language models that bridges the gap between aggregated metrics and fine-grained performance analysis. By integrating contextual embeddings with model probabilities into a unified, performance-aware representation and leveraging a sparse autoencoder to extract human-interpretable features, our approach enables the discovery of coherent data slices where one model outperforms another. Our experiments---spanning variations in model family, size, and post-training regimes---demonstrate that \textsc{BehaviorBox} can uncover nuanced performance differences, such as distinctions in formatting, domain-specific language, and syntactic patterns, that are often masked by conventional evaluation metrics like perplexity. Additionally, our results highlight how differences in perplexity are not correlated with the number of salient features that distinguish models. A delta in perplexity could be a result of noisy, less coherent differences in behavior (as in our comparisons across sizes), while in other cases (e.g. across model families) a minuscule delta could potentially hide a number of well-stratified groups.

Beyond its utility for detailed performance diagnostics, \textsc{BehaviorBox} serves as a hypothesis generation tool for further behavioral analysis and facilitates a deeper understanding of language model behavior, thereby supporting more informed decisions in model development and deployment. Overall, our method represents a step toward more transparent and actionable insights into the inner workings of large-scale language models.

\section{Limitations}

While \textsc{BehaviorBox} shows promise as an interpretability and diagnostic tool, several limitations warrant discussion. First, the approach is dependent on the quality and compatibility of the underlying contextual embeddings and probability estimates. Any misalignment between the embedding space and the performance signals can obscure meaningful differences. Second, aggregating token-level probabilities into word-level metrics may introduce noise, particularly when tokenization strategies differ across models. 

Additionally, the sparse autoencoder, despite its design for interpretability, may not capture all relevant behavioral nuances, and its performance is sensitive to hyperparameter choices such as the sparsity level and the weighting of probability features. The automated labeling process---while efficient---relies on a strong LLM annotator, which can sometimes generate inconsistent or suboptimal descriptions. Finally, our experiments have been conducted on a subset of language modeling tasks and datasets; thus, the generalizability of \textsc{BehaviorBox} to other tasks, domains, or non-textual modalities remains to be fully explored. Future work may address these limitations by refining the representation alignment, exploring alternative aggregation strategies, and broadening the scope of evaluation.

\section*{Ethical Considerations}

\textsc{BehaviorBox} provides new tools for practitioners to better understand the behavior of language models, and particularly the differences between multiple language models.
On the whole, this has the potential for easing the ethical deployment of language models by identifying potential issues in advance of deployment and rectifying them before their deployment.
Overall, we foresee few ethical risks in the existence of such a framework, although as with all automatic tools, users must be cautious in jumping to conclusions based solely on the tool output without careful thought.

\section*{Acknowledgments}
This work was supported in part by a gift from Amazon Web Services. We thank Alex Fang for helping create a visualization interface to inspect features, and Amanda Bertsch, Jared Fernandez, Sireesh Gururaja, Sophie Hao, Michael Hu, Emmy Liu, Jeremiah Milbauer, Lintang Sutawika, Vijay Viswanathan, and members of NeuLab for their feedback and helpful discussion.

\bibliography{custom}

\appendix

\section{Appendix}
\subsection{SAE Hyperparameters and Training}
\label{sec:hyperparams}

\begin{table}[t]
    \centering
    \begin{tabular}{lc} 
    \toprule
    Hyperparameter & Value \\ 
    \midrule
    Batch size & 128 \\
    Learning rate & $10^{-4}$ \\
    AdamW $\beta_1$ & 0.9 \\
    AdamW $\beta_2$ & 0.99 \\
    Dict size & 3000 \\
    $k$ & 50 \\
    Probability feature weight & 0.7 \\
    \bottomrule
    \end{tabular}
    \caption{Hyperparameters used to train SAEs.}
    \label{tab:sae}
\end{table}

Hyperparameters used to train the SAE are included in \autoref{tab:sae}. Hyperparameters were chosen based on a number of heuristics and proxy metrics, including the number of dead latents and the number of resulting coherent features as determined by the LLM annotation process. As there are a large number of hyperparameters and potential values, we did not explore varying all parameters; systematically choosing these parameters has been noted as a challenge in previous work \cite{bricken2023towards}. We explored using both larger dictionaries (e.g. 4000, 5000) and applying greater sparsity constraints ($k=$10, 25), but didn't find a consistent pattern with respect to how these changes affect the number and types of features found across comparisons. 

\paragraph{Probability feature up-weighting} A hyperparameter that had a significant impact on downstream results with a clear trend was the up-weighting of probability features. Using a smaller corpus of about 8M words, we train SAEs where we only vary how much we up-weigh the probabilities, i.e. the fraction of the total magnitude that the probability features contribute. As expected, as this fraction increases, features that are discovered become more consistent with respect to the sign of the probability difference between models for words associated with that feature; however, there is a tradeoff with the coherence of those features (with a high fraction, features are often solely depending on large probability diffs alone, without regards to semantic similarity) and the number of dead latents, as seen in Table \ref{tab:ofw}. As a result, we settled on using a fraction of 0.7 in our experiments. 

\begin{table}[h]
    \centering
    \footnotesize
    \begin{tabular}{cccc} 
    \toprule
    Prob Fraction & \% Dead $\downarrow$ & Word Dist $\downarrow$ & Prob Dist $\downarrow$ \\ 
    \midrule
    0.5 & 14.7 & 8.59 & 0.33 \\
    0.6 & 7.0 & 8.60 & 0.34 \\
    0.7 & \textbf{4.5} & \textbf{8.25} & 0.30 \\
    0.8 & 20.7 & 8.44 & \textbf{0.26} \\
    0.9 & 60.5 & 8.65 & 0.28 \\
    \bottomrule
    \end{tabular}
    \caption{Sweep over the degree probability up-weighting using SAEs trained on 8M words of data (model probabilities used are from Llama 2 7B and 13B). \% dead refers to the number of SAE features that never activate. For each feature, among the (up to) 50 words that activate the feature, we calculate the mean L2 distance between each word's contextual embedding and the average embedding, and do the same for the probability vector.}
    \label{tab:ofw}
\end{table}

\paragraph{Neuron resetting} Additionally, to mitigate the presence of dead latents during training, we follow the methodology in \citet{bricken2023towards} and periodically re-initialize encoder and decoder weights for features that have no non-zero activations on a hold-out eval set during training. For our experiments, we reset dead neurons every 30k steps, but only if the percentage of dead neurons (based on inference on the eval set) is greater than 15\%; the frequency of this resetting and threshold may be adjusted depending on the amount of data available.

\paragraph{What about random seeds?} Works such as \citet{fel2025archetypal} have pointed out that multiple runs on the same data with different random seeds or data orderings can largely impact the space of features found by SAEs. Indeed, in our experiments we also noticed variation (more so with changes in initialization as opposed to data shuffling), though there appeared to be some overlap, especially among either thematically repetitive features (e.g. tabs in varying contexts) or among features with large probability differences. We hypothesize that different runs are capable of finding partially disjoint, yet ``correct'', sets of features, and in practice one could take the union of various SAE runs as their feature set of interest, or use methods such as those proposed in \citet{fel2025archetypal}.

\subsection{Thresholding Features}
\label{sec:thresholding}
Our initial experiments used a two-sided t-test to filter features based on whether difference in means was statistically significant (as opposed to the median difference), but found that this was not suitable as many distributions of differences were not Gaussian. As a result, we encountered features which had statistically significant difference in means (as determined by the t-test) indicating that some model A was better than B, but was very inconsistent with respect to the number of samples where A was actually better than B.

\begin{figure}[!h]
    \centering
    \subfloat{
        {\includegraphics[width=1\linewidth]{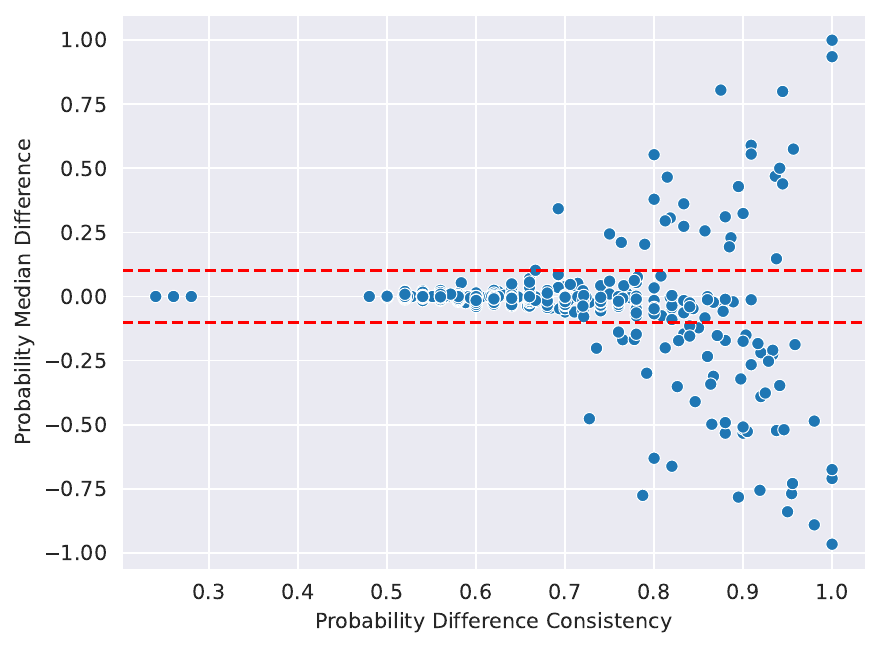}}
    } \\
    \subfloat{
        {\includegraphics[width=1\linewidth]{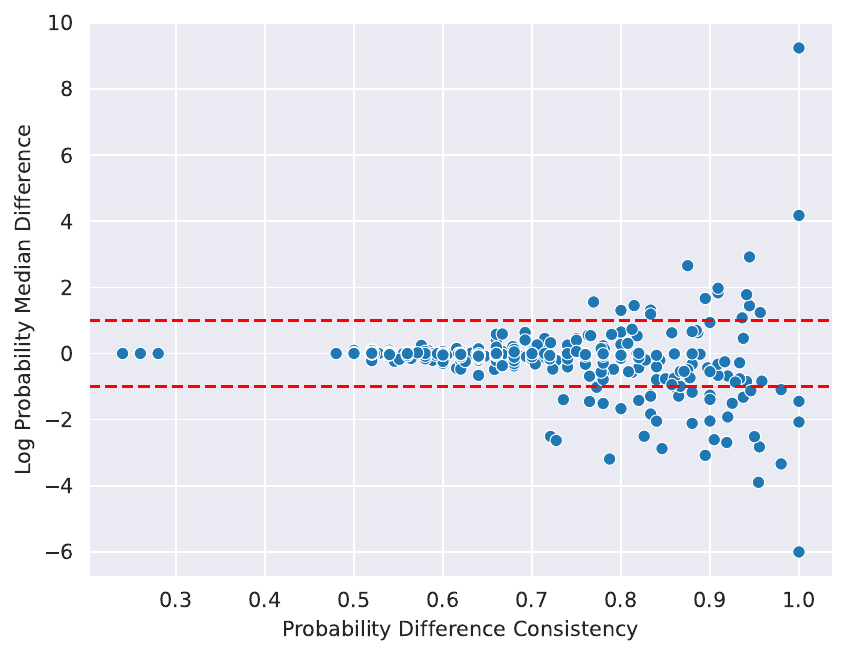}}
    }
    \caption{Scatterplots showing median probability (top) and log probability (bottom) of features (after the first pass annotation step) against consistency. Features are from Llama2-7B and Llama2-13B, and other comparisons generally follow this visual trend. Red lines show our thresholding values ($\pm 0.1$ for probability and $\pm1$ for log probability).}%
    \label{fig:thresholding}
\end{figure}

Thus, we instead use the median difference of the either probabilities or log probabilities as a metric to decide whether to keep or discard a feature for analysis. To choose a threshold for the delta in probabilities and log probabilities, we look at the consistency of a feature, where consistency is defined as the percentage of samples where the sign of the difference between the two model probabilities matches the median difference. As shown in Figure \ref{fig:thresholding}, these chosen thresholds allow us to filter out most features that have a consistency of $<70\%$.

\subsection{LLM Annotation}
\label{sec:prompts}
The first pass annotation prompt (as shown below) was used to generate a first-pass of annotation labels. The top 20 words that lead to the highest activation value (or up to 20 for features containing greater than 10 but less than 20 samples with non-zero activations) and their contexts for a feature were provided in a list following the prefix.

To filter and validate the features, we have an additional round of LLM annotation, which takes as input the original label from the annotator LLM along with (up to) the top 20 words and their contexts. In our qualitative analyses, we only consider the labels output from this labeling stage that were scored $\geq 1$ (as labels scored 1 or 2 were re-labeled). We then manually verify feature labels, and additionally filter out samples that are not described by the label for each feature. \\

\clearpage
\newpage
\begin{minipage}{0.9\textwidth}
    \begin{tcolorbox}[enhanced jigsaw,breakable,pad at break*=1mm,
      colback=blue!5!white,
      colframe=blue!75!black,
      title=First Pass Annotation Prompt,
      fontupper=\normalfont
      ] Your job is to determine if a group of words (surrounded by asterisks, e.g. *word*) in specific contexts form a coherent group that can be described concisely. I will provide you with a list of words surrounded by asterisks and the context in which they appear, usually within a sentence or a block of text. Each word and how it appears in context will be its own item in a list. \\ \\
        Here are some examples: \\
- conservation: Efforts in *conservation* are essential for protecting endangered species. \\
- habitat: The loss of *habitat* is a significant threat to biodiversity. \\
- ecosystem: An *ecosystem* needs a balance of various species to thrive. \\ \\
Your job is to determine if the words form a coherent group that can be described concisely. \
If the words do form a coherent group, please provide a concise description of the group. \
Provide your answer in the following format: \\ \\
<BEGIN ANSWER> \\
Coherent: <YES or NO> \\
Description: <if YES above, your description here; otherwise NONE> \\
<END ANSWER> \\ \\
Do not provide any additional text after <END ANSWER>. \
Only respond with YES or NO for the "Coherent" field. \
If you respond with YES, you must provide a description in the "Description" field. \
Descriptions should be concise, ideally a single sentence. \
For the above example, descriptions may be something like "Nouns describing environmental conservation" or "Words related to biodiversity". \
Note that groups and descriptions may also pertain to formatting, such as "Punctuation before whitespaces in documents discussing logic" or "Series of whitespaces in documents discussing visual art".\
The description should NOT refer to the asterisks, those are only there to help you identify the words. \\ \\
Please categorize the following list of words and their contexts as coherent or not coherent, and provide a description if needed:
    \end{tcolorbox}
\end{minipage}

\vskip\baselineskip

\clearpage
\begin{minipage}{0.9\textwidth}
    \begin{tcolorbox}[enhanced jigsaw,breakable,pad at break*=1mm,
      colback=blue!5!white,
      colframe=blue!75!black,
      title=Label Validation Prompt,
      fontupper=\normalfont
      ] Your job is to determine if a group of words (surrounded by asterisks, e.g. *word*) in specific contexts form a coherent group that is accurately described by a given label. \
I will provide you with a list of words surrounded by asterisks and the context in which they appear, usually within a sentence or a block of text. \
Each word and how it appears in context will be its own item in a list.
Determine if the words form a group that is accurately described by the label by providing a numerical score (0 to 3, and -1).
Scores are defined as follows: \\ \\
- 0: The label is not accurate and the words do not form any coherent groups. \\
- 1: The label is not accurate, but the words form a coherent group. \\
- 2: The label is accurate, but fails to capture a more specific trend. \\
- 3: The label is accurate and captures a specific trend. \\
- -1: There are two coherent groups. \\ \\
Additionally, if you give a score of 1 or 2, provide an alternative label that you believe would be more accurate. \
If you give a label of -1, provide a label for each group. Each label should be separated with <SEP>. \
This label should be precise, concise, and accurate, ideally a single sentence, \
Otherwise, leave the alternative label field blank. \\ \\
Provide your answer in the following format, be sure to include both "Score" and "Label" fields: \\ \\
<BEGIN ANSWER> \\
Score: <a number between 1-3 or -1> \\
Label: <label(s) if original score is 1, 2, or -1, empty otherwise> \\
<END ANSWER> \\ \\
Do not provide any additional text after <END ANSWER>. \
Only respond a number between 0 and 3 or -1 in the Score field. \
The description should NOT refer to the asterisks, those are only there to help you identify the words. \
If there are double asterisks in the text, assume the word of interest is the whitespace between them. \\ \\
Please score the following list of words and their label, and provide a new label if necessary:
    \end{tcolorbox}
\end{minipage}

\vskip\baselineskip
\clearpage

\subsection{Feature Labels}
\label{sec:feature-labels}
Below we list all features for each comparison. Since words that weren't satisfied by the label within each feature were filtered out, the median probability and log probability differences were recalculated

Additionally, we sometimes found repeated features, or features that are activated by a large proportion of the same words. These features tend to occur when the magnitude of the probability difference is relatively large, and there are a large enough nearly identical contexts in which this word occurs within the corpus. 
\begin{table*}[h]
    \centering
    \footnotesize
    \caption{Llama2-7B vs. Llama2-13B}
    \begin{tabular}{>{\centering\arraybackslash}m{0.1\textwidth} p{0.45\textwidth} 
    >{\centering\arraybackslash}m{0.1\textwidth}
    >{\centering\arraybackslash}m{0.1\textwidth}
    >{\centering\arraybackslash}m{0.1\textwidth}
    }
    \toprule
    Model & Feature & Median Prob Diff & Median Log Prob Diff & Consistency \\ 
    \midrule
    Llama2-7B & Tabs following another tab, typically after sentence-ending punctuation marks in narrative text & 0.999 & 9.27 & 1.0 \\
    Llama2-7B & Tabs in narrative text, likely from historical fiction & 0.553 & 1.303 & 0.846 \\
    Llama2-7B & Tabs in primarily narrative text & 0.723 & 1.819 & 0.875 \\
    Llama2-7B & Common English expressions and idioms used in casual dialogue or narrative text & 0.204 & 0.578 & 0.769 \\
    Llama2-7B & References to military related events or figures & 0.03 & 0.213 & 0.818 \\
    \midrule \midrule
    Llama2-13B & Whitespace characters following textual content in book formatting and typography & -0.556 & -2.042 & 0.909 \\
    Llama2-13B & Tabs, commas, and periods in printed formatting & -0.412 & -0.883 & 0.969 \\
    Llama2-13B & Casual dialogue and vernacular speech patterns in literary text & -0.197 & -1.775 & 0.75 \\
    Llama2-13B & Author surname abbreviations followed by parenthetical punctuation in bibliographic entries & -0.143 & -0.29 & 0.787 \\
    Llama2-13B & Informal expressions indicating someone's state of being that include the preposition "to" & -0.013 & -0.797 & 0.778 \\
    Llama2-13B & Book titles beginning with "Or" used as subtitles or alternative titles, typically in publishing contexts & -0.08 & -0.09 & 0.889 \\
    Llama2-13B & Descriptions of landscape features and human interactions with natural and cultivated environments & -0.003 & -0.003 & 0.762 \\
    Llama2-13B & Negative contractions in English informal or vernacular dialogue ("didn't", "wouldn't", "carn't") & -0.0 & -0.585 & 0.667 \\
    \bottomrule
    \end{tabular}
\end{table*}
\begin{table*}[t]
    \centering
    \footnotesize
    \caption{OLMo2-7B vs. OLMo2-13B}
    \begin{tabular}{>{\centering\arraybackslash}m{0.1\textwidth} p{0.45\textwidth} 
    >{\centering\arraybackslash}m{0.1\textwidth}
    >{\centering\arraybackslash}m{0.1\textwidth}
    >{\centering\arraybackslash}m{0.1\textwidth}
    }
    \toprule
    Model & Feature & Median Prob Diff & Median Log Prob Diff & Consistency \\ 
    \midrule
OLMo2-7B & References to original illustrations in HTML file documentation & 0.944 & 3.653 & 1.0 \\
OLMo2-7B & References to original illustrations in HTML document metadata & 0.944 & 3.653 & 1.0 \\
OLMo2-7B & References to original illustrations in HTML file versions across different document numbers & 0.944 & 3.653 & 1.0 \\
OLMo2-7B & Consistent reference to original illustration in file metadata HTML directives & 0.944 & 3.653 & 1.0 \\
OLMo2-7B & File extensions in URLs or file paths indicating compressed archive files & 0.661 & 1.162 & 0.979 \\
\midrule \midrule
OLMo2-13b & Double commas that appear in church registries and marriage records as row separators & -0.627 & -2.082 & 1.0 \\
OLMo2-13b & Double commas that appear in church registries and marriage records as row separators & -0.337 & -0.959 & 0.98 \\
OLMo2-13b & Stage directions indicating character thoughts or private speech in dramatic texts & -0.421 & -0.816 & 1.0 \\
OLMo2-13b & Distribution type indicator in Gradle version file paths & -0.315 & -0.441 & 1.0 \\
OLMo2-13b & Action verbs describing physical or mechanical movement and control & -0.046 & -2.768 & 0.947 \\
OLMo2-13b & Misspelled or phonetically written words in informal or historical texts, specifically "wood" for "would" & -0.032 & -1.554 & 1.0 \\
OLMo2-13b & Ampersand symbols used as abbreviations for "etc." or "and" in various bibliographic and listing contexts & -0.005 & -0.987 & 0.727 \\
OLMo2-13b & Words specifying individual entities or items of a type & -0.003 & -1.427 & 0.81 \\
    \bottomrule
    \end{tabular}
\end{table*}
\begin{table*}[t]
    \centering
    \footnotesize
    \caption{Llama2-7B vs. Llama2-7B-Chat}
    \begin{tabular}{>{\centering\arraybackslash}m{0.14\textwidth} p{0.41\textwidth} 
    >{\centering\arraybackslash}m{0.1\textwidth}
    >{\centering\arraybackslash}m{0.1\textwidth}
    >{\centering\arraybackslash}m{0.1\textwidth}
    }
    \toprule
    Model & Feature & Median Prob Diff & Median Log Prob Diff & Consistency \\ 
    \midrule
    Llama2-7B & References to public domain material in library and archive digitization projects & 0.968 & 3.498 & 1.0 \\
Llama2-7B & Library source attributions in document metadata & 0.952 & 6.541 & 1.0 \\
Llama2-7B & Library source attributions in document metadata & 0.952 & 6.541 & 1.0 \\
Llama2-7B & Source attribution statements referencing public domain materials & 0.968 & 4.696 & 1.0 \\
Llama2-7B & Occurrences of "by" followed by digital libraries, archives, or institutional sources in document attribution statements & 0.952 & 6.541 & 1.0 \\
Llama2-7B & Double commas used as separators in parish register entries containing dates and personal records & 0.726 & 2.536 & 1.0 \\
Llama2-7B & Double commas used as separators in genealogical or parish records between dates and personal information & 0.723 & 2.3 & 1.0 \\
Llama2-7B & Double commas used as separators in genealogical or parish records & 0.635 & 4.001 & 1.0 \\
Llama2-7B & Words indicating location, direction, or movement in narrative text & 0.225 & 2.708 & 1.0 \\
\midrule \midrule
Llama2-7B-Chat & Tabs following a tab preceding dialogue or narrative transitions & -1.0 & -10.989 & 1.0 \\
Llama2-7B-Chat & Word "or" as a URL separator between file path and domain in Project Gutenberg URLs & -0.99 & -4.604 & 1.0 \\
Llama2-7B-Chat & The word "or" appearing in Project Gutenberg URLs as a separator between elements & -0.989 & -4.519 & 1.0 \\
Llama2-7B-Chat & Semicolons appearing in HTML markup and references sections & -0.364 & -0.564 & 1.0 \\
Llama2-7B-Chat & Imperative commands using "with" to indicate dismissal or removal & -0.295 & -0.446 & 1.0 \\
Llama2-7B-Chat & Words and phrases indicating decision-making, judgment, or reaching conclusions in formal or narrative contexts & -0.156 & -0.718 & 0.844 \\
Llama2-7B-Chat & Instances of "with a" followed by words describing facial expressions or emotional gestures in narrative text & -0.168 & -0.227 & 0.939 \\
Llama2-7B-Chat & Commas followed by forms of address or names in direct dialogue & -0.15 & -0.265 & 0.84 \\
Llama2-7B-Chat & Conditional uses of 'were' in hypothetical scenarios, typically following 'if you' & -0.14 & -0.427 & 0.76 \\
Llama2-7B-Chat & Instances of the word "reached" used to describe arriving at or attaining a destination in narrative texts & -0.19 & -0.385 & 0.744 \\
Llama2-7B-Chat & Superlative adjective "most" used to express highest degree or maximum quality of something & -0.132 & -0.286 & 0.76 \\
Llama2-7B-Chat & Uses of "who" as a relative pronoun referring to people or beings in formal or literary contexts & -0.109 & -0.187 & 0.94 \\
Llama2-7B-Chat & Commas preceeding "and" in various narrative and instructional texts & -0.128 & -0.23 & 0.8 \\
Llama2-7B-Chat & Exclamation marks appearing at the end of emotionally charged or dramatic statements & -0.103 & -0.218 & 0.86 \\
Llama2-7B-Chat & Modal verb 'must' expressing necessity, obligation, or requirement in various contexts & -0.129 & -0.184 & 0.826 \\
Llama2-7B-Chat & Forms of the verbs "to be" (were/are) used as auxiliary or linking verbs in historical or narrative contexts & -0.103 & -0.161 & 0.72 \\
Llama2-7B-Chat & Informal or dialectical spelling of "and" in literary texts showing vernacular speech & -0.114 & -0.228 & 0.66 \\
    \bottomrule
    \end{tabular}
\end{table*}
\begin{table*}[t]
    \centering
    \footnotesize
    \caption{OLMo2-7B vs. OLMo2-7B-DPO}
    \begin{tabular}{>{\centering\arraybackslash}m{0.14\textwidth} p{0.41\textwidth} 
    >{\centering\arraybackslash}m{0.1\textwidth}
    >{\centering\arraybackslash}m{0.1\textwidth}
    >{\centering\arraybackslash}m{0.1\textwidth}
    }
    \toprule
    Model & Feature & Median Prob Diff & Median Log Prob Diff & Consistency \\ 
    \midrule
OLMo2-7B & The word "genus" appearing in taxonomic descriptions of ray-finned fish species belonging to Enteromius & 0.715 & 3.876 & 1.0 \\
OLMo2-7B & Words related to influence, vulnerability, persuasiveness, and power in conflict & 0.08 & 1.154 & 0.933 \\
OLMo2-7B & Instances of the word "said" used as a dialogue tag in narrative fiction & 0.181 & 0.433 & 0.9 \\
OLMo2-7B & Instances of "Very Young Man" as a character name in a narrative text & 0.155 & 0.824 & 0.796 \\
OLMo2-7B & Words expressing challenges, obstacles or removal in narrative contexts & 0.09 & 2.065 & 0.917 \\
OLMo2-7B & Action words or numbers serving as interactive elements or references in digital/printed content & 0.04 & 1.769 & 0.893 \\
OLMo2-7B & "Place" as used in the phrase "place of" or as a rank & 0.008 & 2.524 & 0.909 \\
\midrule \midrule
OLMo2-7B-DPO & Instances of "so much the" followed by comparative expressions in dialogue & -0.584 & -1.138 & 1.0 \\
OLMo2-7B-DPO & Requests for private conversations using the phrase "have a word with" & -0.239 & -0.307 & 1.0 \\
    \bottomrule
    \end{tabular}
\end{table*}
\onecolumn
\footnotesize
\begin{longtable}
{>{\centering\arraybackslash}m{0.12\textwidth} p{0.43\textwidth} 
    >{\centering\arraybackslash}m{0.1\textwidth}
    >{\centering\arraybackslash}m{0.1\textwidth}
    >{\centering\arraybackslash}m{0.1\textwidth}}
    \caption{Llama2-7B vs. OLMo2-7B} \\
    \toprule
    Model & Feature & Median Prob Diff & Median Log Prob Diff & Consistency \\ 
    \midrule
    \endfirsthead
    Llama2-7B & Tabs used for formatting in all-caps chapter titles and section headers & 0.995 & 5.471 & 1.0 \\
Llama2-7B & References to HTML file versions in Project Gutenberg notes & 0.888 & 2.205 & 1.0 \\
Llama2-7B & References to HTML file format in Project Gutenberg documentation notes & 0.843 & 1.875 & 1.0 \\
Llama2-7B & References to HTML versions of files in Project Gutenberg documentation & 0.843 & 1.875 & 0.98 \\
Llama2-7B & Words indicating spatial or temporal orientation & 0.822 & 2.687 & 0.917 \\
Llama2-7B & Imperative "See" commands directing users to HTML files containing illustrations & 0.677 & 1.137 & 1.0 \\
Llama2-7B & Opening quotation marks in dialogue appearing after two tabs & 0.657 & 1.359 & 1.0 \\
Llama2-7B & Opening quotation marks in dialogue appearing after two tabs & 0.635 & 1.399 & 1.0 \\
Llama2-7B & Periods appearing in Java-style package and class references and application configuration settings & 0.629 & 1.03 & 1.0 \\
Llama2-7B & Opening quotation marks in dialogue appearing after two tabs & 0.522 & 1.193 & 1.0 \\
Llama2-7B & Tabs following semicolons in poetry written in nonstandard English dialects or other languages & 0.456 & 0.657 & 1.0 \\
Llama2-7B & Tabs following tabs and preceding quotation marks marking direct speech in narrative text & 0.402 & 0.535 & 0.96 \\
Llama2-7B & Tabs following tabs in narrative text & 0.375 & 0.47 & 0.857 \\
Llama2-7B & Tabs after puncutation (commas, semicolons) and before the word "and" in literary passages & 0.34 & 0.482 & 0.92 \\
Llama2-7B & Tabs followed by another tab between segments of dialogue in narrative text & 0.305 & 0.364 & 0.694 \\
Llama2-7B & Tabs in multilingual software configuration and documentation & 0.29 & 0.43 & 1.0 \\
Llama2-7B & Tabs in segments of dialog showing character reactions or responses following quoted speech & 0.288 & 0.34 & 1.0 \\
Llama2-7B & Dialectical and informal variations of words, particularly showing dropped 'h' sounds & 0.237 & 0.569 & 0.8 \\
Llama2-7B & Tabs following tabs between quoted text between two speakers & 0.224 & 0.254 & 1.0 \\
Llama2-7B & Tabs within whitespace sequences in dialogue segments in narrative texts & 0.223 & 0.278 & 0.9 \\
Llama2-7B & Tabs followed by another tab before quotations in literary dialogue & 0.201 & 0.232 & 0.688 \\
Llama2-7B & Opening quotation marks at the start of dialogue or exclamations in narrative text & 0.189 & 0.504 & 0.84 \\
Llama2-7B & Tabs followed by another tab following quotations in literary dialogue & 0.187 & 0.29 & 0.776 \\
Llama2-7B & Single or double tabs appearing between chapters, sections, or bibliographic entries in a historical text & 0.167 & 0.343 & 0.9 \\
Llama2-7B & Question marks followed by quotation marks at the end of dialogue in literary texts & 0.147 & 0.308 & 0.857 \\
Llama2-7B & Sequences of tabs after in book layout and formatting contexts & 0.142 & 0.224 & 0.86 \\
Llama2-7B & Periods at the end of sentences, including contexts from online conversations and comments & 0.134 & 0.264 & 0.75 \\
Llama2-7B & Whitespace (spaces, tabs) in poetry & 0.12 & 0.254 & 0.7 \\
Llama2-7B & Opening quotation marks and brackets at the start of bibliographic or reference entries & 0.052 & 0.06 & 0.559 \\
\midrule \midrule
OLMo2-7B & HTML "less than" symbols used as line break tags in geographic reference texts & -0.986 & -4.291 & 1.0 \\
OLMo2-7B & Roman numerals VIII and XIII appearing as sequential chapter or section numbers in document structure & -0.952 & -4.493 & 1.0 \\
OLMo2-7B & Numerical and textual elements appearing in index entries, lists, and content references & -0.936 & -6.076 & 0.909 \\
OLMo2-7B & Occurrences of the word "wich" in parenthetical titles and signatures by Petroleum V. Nasby & -0.891 & -9.468 & 1.0 \\
OLMo2-7B & Items appearing as numerical markers or identifiers in ordered lists or sections<SEP> & -0.818 & -2.762 & 1.0 \\
OLMo2-7B & Whitespace between sentence boundaries in narrative texts & -0.608 & -0.94 & 1.0 \\
OLMo2-7B & Periods followed by quotation marks before another quote in narrative dialogue & -0.326 & -0.797 & 0.96 \\
OLMo2-7B & Government and legislative bodies, particularly British Parliament and colonial institutions & -0.318 & -2.124 & 0.952 \\
OLMo2-7B & Question-ending punctuation marks followed by quotation marks in dialogue, often with rhetorical or interrogative phrases ending in "eh" or similar words & -0.183 & -0.277 & 0.875 \\
OLMo2-7B & Whitespace after dialogue tags and before quoted continuations in narrative dialogue & -0.16 & -0.175 & 1.0 \\
OLMo2-7B & Commas preceding tabs and narrative continuation or descriptive elaboration in poetic text & -0.154 & -0.279 & 0.94 \\
OLMo2-7B & Whitespace between sentences in formatted literary text & -0.147 & -0.159 & 1.0 \\
OLMo2-7B & Punctuation marks at the end of poetic or literary lines & -0.129 & -0.171 & 0.82 \\
OLMo2-7B & Tabs followed by another tab and dialog or quoted speech in narrative text & -0.126 & -0.192 & 0.673 \\
OLMo2-7B & Semicolons used as separators in index or reference entries & -0.111 & -0.146 & 0.92 \\
OLMo2-7B & Whitespace between sentenes in philosophical and sociological texts & -0.102 & -0.109 & 0.96 \\
    \bottomrule
%     \end{tabular}
% \end{table*}
\end{longtable}
\twocolumn
\onecolumn
\footnotesize
\begin{longtable}
{>{\centering\arraybackslash}m{0.12\textwidth} p{0.43\textwidth} 
    >{\centering\arraybackslash}m{0.1\textwidth}
    >{\centering\arraybackslash}m{0.1\textwidth}
    >{\centering\arraybackslash}m{0.1\textwidth}}
    \caption{Llama2-13B vs. OLMo2-13B} \\
    \toprule
    Model & Feature & Median Prob Diff & Median Log Prob Diff & Consistency \\ 
    \midrule
    \endfirsthead
    Llama2-13B & Tabs following another tab in chapter titles or section headers  & 0.989 & 4.567 & 1.0 \\
Llama2-13B & "https" within Project Gutenberg file directory URLs & 0.984 & 6.801 & 1.0 \\
Llama2-13B & "morrow" as part of archaic hyphenated spelling in the word "to-morrow" & 0.756 & 2.519 & 1.0 \\
Llama2-13B & Opening quotation marks followed by various text segments at the start of dialogue or quoted passages & 0.725 & 1.431 & 0.98 \\
Llama2-13B & Opening quotation marks following dialogue consisting of questions in narrative text & 0.666 & 1.328 & 0.98 \\
Llama2-13B & Captions or labels for figures and illustrations in instructional or historical texts & 0.558 & 0.837 & 1.0 \\
Llama2-13B & Opening quotation marks at the start of questions in dialogue & 0.514 & 1.124 & 0.98 \\
Llama2-13B & Tabs following semicolons in poetry written in nonstandard English dialects & 0.41 & 0.546 & 0.96 \\
Llama2-13B & Tabs following another tab between sentences in literary text excerpts & 0.346 & 0.424 & 0.92 \\
Llama2-13B & Tabs preceding another tab between quotes in dialogue between two speakers & 0.288 & 0.341 & 0.896 \\
Llama2-13B & Tabs following another tab after colons and preceding quoted dialogue in narrative text & 0.263 & 0.305 & 1.0 \\
Llama2-13B & Tabs following commas and appearing before capitalized conjunctions & 0.251 & 0.319 & 0.94 \\
Llama2-13B & Tabs following another tab between quotes in dialogue between two speakers & 0.244 & 0.28 & 1.0 \\
Llama2-13B & Quotation marks at the start of exclamations in narrative text & 0.238 & 0.77 & 0.9 \\
Llama2-13B & Tabs preceding another tab between quotes in dialogue between two speakers & 0.236 & 0.289 & 0.86 \\
Llama2-13B & Equal signs appearing in text separators or boundaries between different sections or concepts & 0.15 & 0.186 & 0.96 \\
Llama2-13B & Underscores appearing at line beginnings in dictionary or reference entries & 0.119 & 0.14 & 0.82 \\
Llama2-13B & Tabs typically preceding another tab along with quoted dialogue & 0.107 & 0.212 & 0.755 \\
Llama2-13B & Tabs followed by character names in theatrical script & 0.102 & 0.115 & 0.72 \\
\midrule \midrule
OLMo2-13B & Path component referring to the Gradle wrapper directory in build configuration files & -0.987 & -6.013 & 1.0 \\
OLMo2-13B & References to the Gradle wrapper directory path in build configuration files & -0.987 & -6.013 & 1.0 \\
OLMo2-13B & Path component referring to Gradle wrapper directory in build configuration files & -0.987 & -6.013 & 1.0 \\
OLMo2-13B & Path component 'wrapper' in Gradle build configuration files & -0.987 & -6.013 & 1.0 \\
OLMo2-13B & HTML line break tags in a references section & -0.972 & -3.567 & 1.0 \\
OLMo2-13B & HTML line break tags in a References section & -0.971 & -3.553 & 1.0 \\
OLMo2-13B & HTML line break tags in a References section & -0.97 & -3.494 & 1.0 \\
OLMo2-13B & HTML less-than symbols appearing in reference formatting & -0.965 & -3.343 & 1.0 \\
OLMo2-13B & Periods followed by quotation marks at the end of a quote in dialogue & -0.303 & -0.638 & 0.898 \\
OLMo2-13B & Whitespace following abrupt ends, interruptions, trailing thoughts, or hesitations in speech & -0.264 & -0.943 & 0.68 \\
OLMo2-13B & Whitespace between sentences in formatted literary text & -0.223 & -0.253 & 1.0 \\
OLMo2-13B & Semicolons used as separators in an index or table of contents & -0.198 & -0.282 & 0.92 \\
OLMo2-13B & Double-asterisk markers appearing after quoted dialogue, followed by additional text & -0.181 & -0.201 & 1.0 \\
OLMo2-13B & Commas used as clause separators in narrative or descriptive text & -0.156 & -0.289 & 0.82 \\
OLMo2-13B & Whitespace between sentences in narrative text excerpts & -0.146 & -0.157 & 1.0 \\
OLMo2-13B & Question marks and quotation marks at the end of quoted dialogue or interrogative statements & -0.139 & -0.168 & 0.955 \\
OLMo2-13B & Tabs preceeding another tab in formatted literary text & -0.116 & -0.147 & 0.66 \\
    \bottomrule
%     \end{tabular}
% \end{table*}
\end{longtable}
\twocolumn

\end{document}